\documentclass[letterpaper, 10 pt, conference]{ieeeconf}  % Comment this line out if you need a4paper

\IEEEoverridecommandlockouts                              % This command is only needed if 
\usepackage{graphics} % for pdf, bitmapped graphics files
\usepackage{epsfig} % for postscript graphics files
\usepackage{times} % assumes new font selection scheme installed
\usepackage{amsmath} % assumes amsmath package installed
\usepackage{amssymb}  % assumes amsmath package installed

\let\labelindent\relax
\usepackage{amsthm}  % assumes amsmath package installed
\usepackage{macrodef}
\usepackage[top=0.8in,bottom=1in,left=0.667in,right=0.667in]{geometry}

\makeatletter
\let\NAT@parse\undefined
\makeatother
\usepackage[hidelinks]{hyperref}
\usepackage{cleveref}

\title{\LARGE \bf Riemannian Variational Calculus: \\ Optimal Trajectories Under Inertia, Gravity, and Drag Effects}

\author{Jinwoo Choi, Alejandro Cabrera, and Ross L. Hatton
    \thanks{Jinwoo Choi and Ross L. Hatton are with the Collaborative Robotics and Intelligent Systems (CoRIS) Institute at Oregon State University, Corvallis, OR USA. \{\href{mailto:choijinw@oregonstate.edu}{choijinw}, \href{mailto:Ross.Hatton@oregonstate.edu}{Ross.Hatton}\}@oregonstate.edu}
    \thanks{Alejandro Cabrera is with Instituto de Matem{\'a}tica, Universidade Federal do Rio de Janeiro, Rio de Janeiro, Brazil. \href{mailto:alejandro@matematica.ufrj.br}{alejandro@matematica.ufrj.br}}
    \thanks{We acknowledge the support in part by the Office of Naval Research under awards N00014-23-1-2171 and the Brazilian agency grants CNPq 309847/2021-4, 402320/2023-9 and FAPERJ JCNE E-26/203.262/2017.}
    \thanks{(Corresponding author: Ross L. Hatton.)}
}% <-this % stops a space

\newcommand{\work}{W}

\newcommand{\grad}[1][]{\text{grad}_{#1}}
\newcommand{\hess}[1][]{\text{Hess}_{#1}}

\newcommand{\vbasis}[1]{\partial_{\config^{#1}}}
\newcommand{\cobasis}[1]{\partial\config^{#1}}

\newtheorem{theorem}{Theorem}
\newtheorem{proposition}[theorem]{Proposition}

\crefname{equation}{}{}
\crefname{figure}{Fig}{Figs}

\begin{document}
\crefname{figure}{Fig.}{Figs.}

\maketitle
\thispagestyle{empty}
\pagestyle{empty}

%%%%%%%%%%%%%%%%%%%%%%%%%%%%%%%%%%%%%%%%%%%%%%%%%%%%%%%%%%%%%%%%%%%%%%%%%%%%%%%%
\begin{abstract}
Robotic motion optimization often focuses on task-specific solutions, overlooking fundamental motion principles. Building on Riemannian geometry and the calculus of variations (often appearing as indirect methods of optimal control), we derive an optimal control equation that expresses general forces as functions of configuration and velocity, revealing how inertia, gravity, and drag shape optimal trajectories. Our analysis identifies three key effects: (i) curvature effects of inertia manifold, (ii) curvature effects of potential field, and (iii) shortening effects from resistive force. We validate our approach on a two-link manipulator and a UR5, demonstrating a unified geometric framework for understanding optimal trajectories beyond geodesic-based planning.
\end{abstract}

%%%%%%%%%%%%%%%%%%%%%%%%%%%%%%%%%%%%%%%%%%%%%%%%%%%%%%%%%%%%%%%%%%%%%%%%%%%%%%%%
\section{INTRODUCTION}

Recent advances in computational power and numerical control algorithms have enabled robots (or machines) to execute complex, precise motions across a wide range of applications. In particular, numerical trajectory optimization methods (appearing as direct methods of optimal control~\cite{wensing_optimization-based_2024}) are widely used to generate task-specific solutions, and learning-based control methods~\cite{green_learning_2021} allow robots to generate efficient motions without explicit models by learning from input-output data. However, these approaches often focus on optimizing individual motion strategies without explicitly characterizing the fundamental motion principles—such as how optimal motion strategies adapt to external forces or kinetic energy changes—thereby hindering the development of more generalizable and interpretable motion strategies.

On the other hand, the calculus of variations (appearing as indirect methods of optimal control~\cite{hussein_optimal_2004,hussein_optimal_2008,bloch_nonholonomic_2010}) provides a systematic framework for understanding how small variations in trajectories affect optimality. In Riemannian geometry, it extends to describe a system’s optimal dynamics, by setting a Riemannian metric encoding the system’s kinetic energy or power dissipated by environments~\cite{bullo_tracking_1999, cortes_monforte_geometric_2002, bullo_geometric_2005, ramasamy_geometry_2019}. In this setting, geodesics—the shortest or energy-optimal paths on the manifold—naturally emerge as optimal motion strategies, governed by the geodesic equation, a system of differential equations that characterizes these paths.

\begin{figure}
    \centering
    \includegraphics[width=\linewidth]{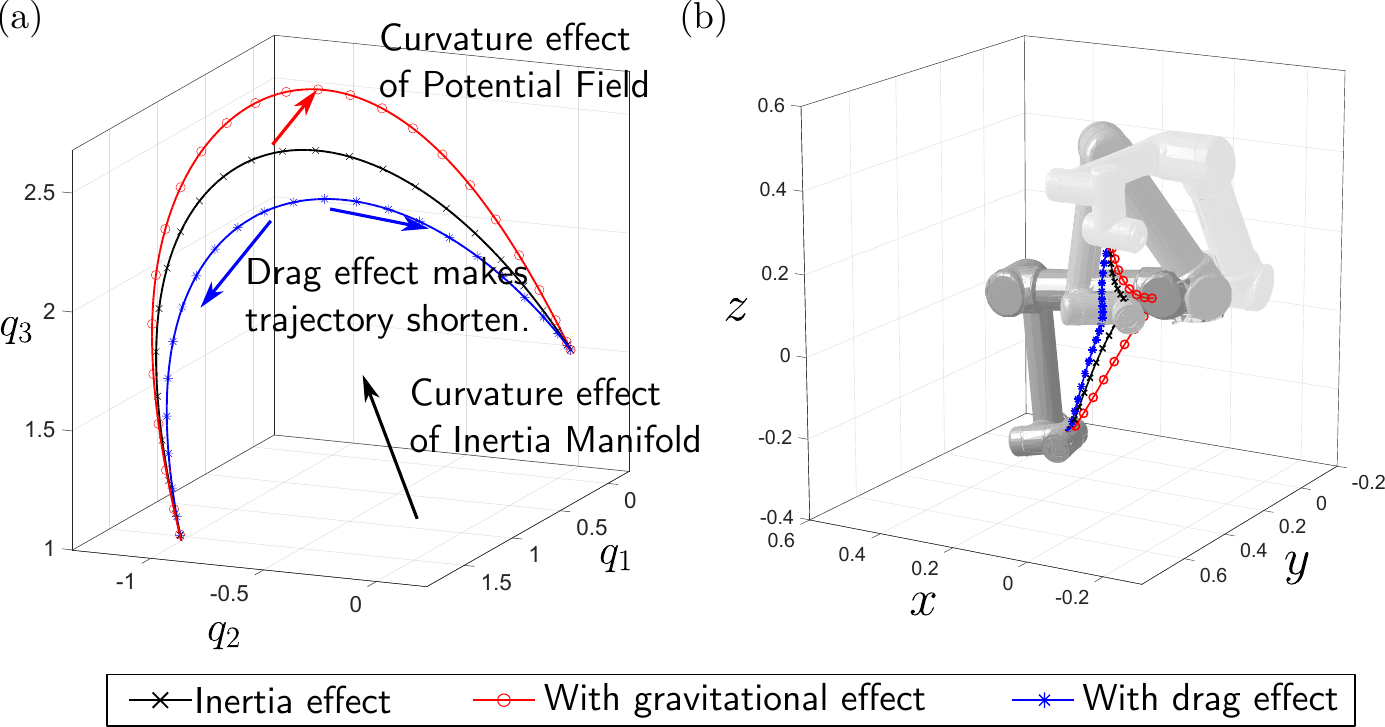}
    \caption{(a) Optimal trajectories for a UR5 manipulator moving its first three joints. The black-$\times$-marked curve considers only inertia effects, where curvature-induced distortion directs acceleration to reduce inertia (e.g., folding the arm). The red-$\circ$-marked curve incorporates both inertia and gravity, influenced by the potential field curvature on the inertia manifold. The blue curve accounts for inertia and drag, where drag, modeled as joint viscous friction, results in an Euclidean drag metric, shortening the trajectory in configuration space. (b) The same optimal trajectories in task space, with the initial and final configurations of the UR5. Only the first three joints, starting from the base joint, are actuated in this work. The manipulator's appearance transitions from white to gray, representing its initial and final configurations.}
    \label{fig:graphical_abstract}
\end{figure}

Although the geodesic equation serves as the optimal control equation for kinematic systems and drag-dominated systems~\cite{ramasamy_geometry_2019}, it does not account for second-order dynamics. For mechanical systems without friction, previous works~\cite{hussein_optimal_2004, hussein_optimal_2008, dubois_pontryagin_2015, balseiro_about_2017} have introduced the Riemannian spline equation to characterize natural (optimal) motion by minimizing acceleration. Additionally, the influence of potential fields on the Riemannian spline equation has been considered~\cite{hussein_optimal_2004,dubois_pontryagin_2015}. Our recent studies~\cite{cabrera_optimal_2024, choi_optimal_2024} introduce an actuation-based cometric and propose a biased Riemannian spline equation to minimize actuation effort. However, to the best of our knowledge, the geometric understanding of optimal trajectories considering the combined effects of kinetic, drag, and gravitational forces remains largely unexplored.

In this paper, building on previous works on Riemannian splines~\cite{balseiro_about_2017, cabrera_optimal_2024, choi_optimal_2024}, we provide a geometric interpretation of optimal trajectories under the combined effects of general forces, in which a general force is expressed as a function of configuration (e.g., joint angles or shape) and velocity. As illustrated in \cref{fig:graphical_abstract}, three key elements shape the structure of optimal trajectories:
\begin{enumerate} 
\item Inertial Effect: Optimal trajectories are influenced by the curvature of the inertia manifold. Generally, positive curvature tends to make trajectories converge at specific points~\cite{cabrera_optimal_2024}.
% To minimize acceleration- or actuation-based costs, trajectories on positively curved manifolds generate acceleration inward.
\item Potential Force Effect: Similar to inertia effects, optimal trajectories are affected by the curvature of potential manifolds, which are linked to inertia manifolds~\cite{hussein_optimal_2004, dubois_pontryagin_2015}. These manifolds represent graphs of potential energy over inertia manifolds.
\item Resistive Force Effect: This effect aligns optimal trajectories with drag manifolds. To minimize energy dissipation from resistive forces, such as joint friction or end-effector drag, it is preferable to follow the shortest path on drag manifolds~\cite{ramasamy_geometry_2019}.
\end{enumerate}

To demonstrate our approach and provide a geometric interpretation, we apply the optimal control equation to a two-link manipulator and a UR5 manipulator, analyzing individual effects in \cref{sec:geomeaning} and their combined influence in \cref{sec:results}. The two-link manipulator, with two joint variables, effectively illustrates key manifolds such as potential fields and drag manifolds. The UR5, a widely used 6-DOF robotic manipulator in research and industry~\cite{kebria_kinematic_2016}, has its first three joints governing end-effector position and the last three controlling orientation. To emphasize inertia and gravity effects, we focus on optimal trajectories in the reduced configuration space of the first three joints.

\section{BACKGROUND}
\subsection{Riemmanian geometry}
In this section, we briefly review \emph{Riemannian Geometry} from the perspective of optimal control problems and mechanical systems. The detailed formulations or definitions of notions can be found in~\cite{lee_riemannian_1997}.

A locally differentiable manifold $\configspace$ equipped with a metric tensor $\gmtensor$ is called a \emph{Riemannian manifold} $(\configspace,\gmtensor)$. Let $\vecbasis \in \euclid^d$ be an open subset with coordinates $\config = (\config^1,\cdots,\config^d).$ $\gmtensor(q)$ is a \emph{Riemannian metric} for the space contains the local distance information. $\vecbasis$ is a coordinate domain in a configuration manifold $\configspace$, and it allows to express explicit coordinate formulas for the tensors. ($\vbasis{i} \equiv \parderiv{}{\config^i}$) denotes the basis of $\tans\vecbasis$ given by (constant) vector fields in $\vecbasis$ and ($\cobasis{i}$) the dual basis of $\tans^*\vecbasis$.  

A tensor is a mathematical object that generalizes scalars, vectors, and matrices to higher dimensions while maintaining certain transformation properties under coordinate changes. A tensor of type $(m, n)$ is a multilinear map that takes in $m$ covectors (covariant components) and $n$ vectors (contravariant components) and produces a scalar~\cite{lee_riemannian_1997}. In coordinates, it is typically represented by a multi-indexed array of numbers.\footnote{We use an Einstein summation convention. A lower index indicates covector elements and an upper index indicates vector elements.} A (tangent) vector on a manifold, as a $(1, 0)$-tensor, describes the rate at which a point moves through a Riemannian manifold $\configspace$. A covector, as a $(0, 1)$-tensor, describes the rate where a quantity changes across a space at a given point. There is a natural product between a covector and a vector because a covector acts as a linear map from tangent vectors to scalars. It is called a covector-vector product,
\begin{equation}
    \deriv{\work}{t} = \covprod{\force}{\dconfig} = \sum_i\force_i\dconfig^i,
\end{equation}
where $\covprod{\cdot}{\cdot}$ denotes a covector-vector product, $\force \in \tans[\config]^*\configspace$ is a covector, $\dconfig \in \tans[\config]\configspace$ is a configuration velocity, and $\work$ is a corresponding scalar quantity. From the viewpoint of mechanics, force fields $\force$ can be interpreted as covector fields, and velocity fields $\dconfig$ can be interpreted as vector fields. Then, a covector-vector product between forces and velocities calculates the time-derivative of work scalars (i.e.,\ power).

There is a natural isomorphism between the cotangent and tangent bundles of Riemannian manifolds via the metric $\gmtensor$:
\begin{subequations}
\begin{align}
   \innerprod{\force}{\bullet}_{\gmtensor^*} &= \gmtensor^{ij}\force_j \vbasis{i} = \acc, 
   \label{eq:mussharp}\\
   \innerprod{\bullet}{\acc}_{\gmtensor} &= \gmtensor_{ij}\acc^i\cobasis{j} = \force,
   \label{eq:musflat}
\end{align}
\end{subequations}
where $\innerprod{\cdot}{\cdot}_{\gmtensor}$ is the inner product with respect to $\gmtensor$, $\force \in \tans[\config]^*\configspace$ is a covector, $\acc \in \tans[\config]\configspace$ is a tangent vector, the bullet notations $\bullet$ represent an element of $\tans[\config]\configspace$ or $\tans[\config]^*\configspace$ so that the corresponding linear function is a dual element in $\tans[\config]^*\configspace$ or $\tans[\config]\configspace$, $\gmtensor_{ij}$ are the components of $\gmtensor$, and $\gmtensor^{ij}$ are the components of the metric inverse (i.e., dual of metric) $\gmtensor^*$, respectively.  The operations in \cref{eq:mussharp} and \cref{eq:musflat} raise and lower vector indices, transforming between vectors and covectors.\footnote{These operations are known as a musical isomorphism operator. For raising an index, a musical sharp symbol is used ($\acc = \force^{\sharp}$). For lowering an index, a musical flat symbol is used ($\force = \acc^{\flat}$).} From the viewpoint of mechanics, raising the index transforms a force covector into an acceleration vector through the mass matrix $\gmtensor(\config)$~\cite{altafini_geometric_2001}.

A covariant derivative is a way to differentiate vector fields (or tensor fields) on a curved space or manifold while respecting the manifold’s geometry. Unlike the standard derivative, which does not account for changes in the coordinate basis, the covariant derivative subtracts out the ``expected changes" in the fields induced by ``warping" of the coordinate chart with respect to the manifold, and so ensures that the result only includes ``true changes" to the field across the manifold. Given a smooth manifold $\configspace$ with a Levi-Civita connection $\riemconn$ and its dual connection $\riemconn^*$, a covariant derivative of a vector $\config$ and a covector $\force$ along the curve $\config(t)$ can be expressed in natural coordinates induced by coordinates $(\config^i)$ in $\configspace$:
\begin{align}
    \left(\covderiv{\dconfig}\dconfig\right)^i &= \ddconfig^i + \ctfl{i}{j}{k}\dconfig^j\dconfig^k, 
    \label{eq:coveccovddef} \\
    \left(\covderiv{\dconfig}^*\force\right)_i &= \dot\force_i - \ctfl{k}{i}{j}\force_k\dconfig^j, 
    \label{eq:veccovdef}
\end{align}
where $\config(t)$ is a time-parameterized curve, and $\ctfl{i}{j}{k}$ is Christoffel symbols~\cite{lee_riemannian_1997}. A \emph{covariant acceleration} $\covderiv{\dconfig}\dconfig$ is a covariant time-derivative of a velocity $\dconfig$ along a curve $\config(t)$. If a covariant acceleration is zero, $\config(t)$ is called a \emph{geodesic}.

A Riemannian curvature tensor $\rcurv \in \Omega^2(\configspace, End(\tans\configspace))$ (see~\cite{balseiro_about_2017}) describes a curvature of the curvature of Riemannian manifolds and its coefficient function $\rcurv^l_{ijk}(\config)$ is defined by
\begin{equation}
    \rcurv^l_{ijk}(\config) = \vbasis{i}\ctfl{l}{j}{k} - \vbasis{j}\ctfl{l}{j}{k} + \ctfl{n}{j}{k}\ctfl{l}{i}{n} - \ctfl{n}{i}{k}\ctfl{l}{j}{n}.
\end{equation}

\section{Optimal Control Problem}

\subsection{Optimal Control System}
The optimal control system is defined by the data ($\configspace, \gmtensor, \inmtensor, \extforce$), where $\configspace$ is configuration manifold, $\gmtensor$ is a ``mass'' metric, $\inmtensor$ is an induced metric encoding control cost~\cite{cabrera_optimal_2024}, and $\extforce$ is an external force dependent on configuration and velocity, including gravitational\footnote{Although a Lagrangian includes potential energy, we interpret a gravitational force as an external force to ``inertia manifold'' in this paper.} and drag forces.

The system is defined as follows:
\begin{subequations}
\begin{align}
\text{State:} \ & \constate(t) = (\config(t),\dconfig(t)) \in \tans\configspace, \\
\text{Control:} \ & \covderiv{\dconfig}\dconfig = \coninput + \extacc(\config,\dconfig), \label{eq:coneq} \\
\text{Cost:} \ & \int_{t_0}^{t_f}\norm{\coninput}^2_{\inmtensor}dt, \label{eq:concost} \\
\text{Initial Condition:} \ & \config(t_0) = \config_0, \ \ \dconfig(t_0) = \vel_0, \\
\text{Final Condition:} \ & \config(t_f) = \config_f, \ \dconfig(t_f) = \vel_f, 
\end{align}
\end{subequations} where $\coninput$ is a control input, and $t_0$ and $t_f$ denote the initial and final times of trajectory, respectively. The term $\extacc = \innerprod{\extforce}{\bullet}_{\gmtensor^*}$ represents the acceleration field induced by external force $\extforce$. Note that the control equation \eqref{eq:coneq} corresponds to the system's equation of motion where the mass matrix is given by the metric tensor $\gmtensor$. The interpretation of the cost functional depends on the induced metric. If the induced metric coincides with the dual metric, the cost functional represents an acceleration-based input norm. Given the actuation-based cometric, the induced metric and the actuator-torque norm cost can be defined as 
\begin{equation}
\inmtensor = \transpose{\gmtensor}\cogmtensor\gmtensor, \quad \norm{\coninput}^2_{\inmtensor} = \norm{\force}_{\cogmtensor}^2,
\end{equation}
with $\cogmtensor$ being the actuation-based cometric tensor, which simplifies to the Euclidean metric when joint and actuator configurations are identical (see~\cite{cabrera_optimal_2024,choi_optimal_2024}).

\subsection{Optimal Spline Equation}
\begin{figure}
    \centering
    \includegraphics[width=\columnwidth]{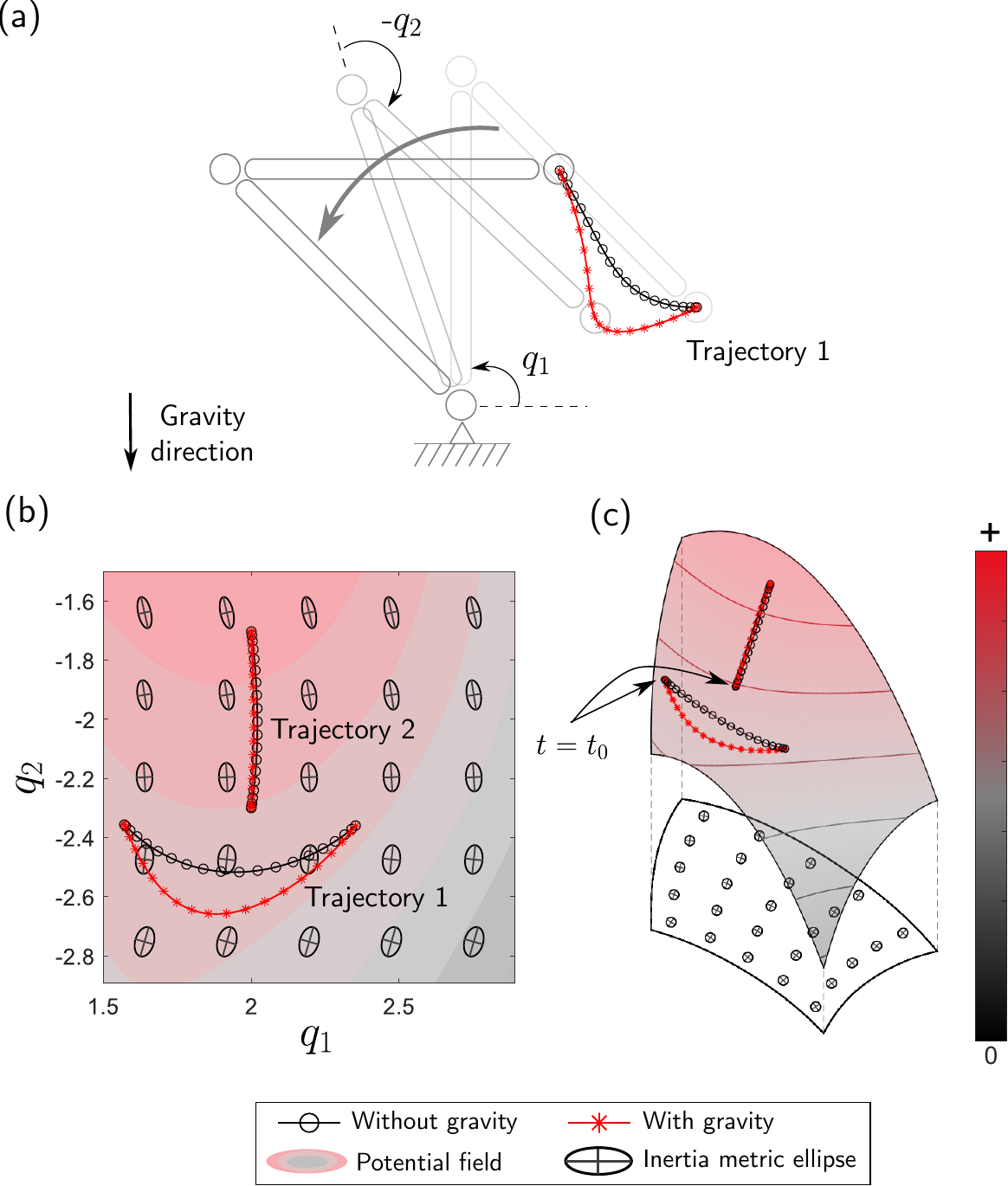}
    \caption{(a) A two-link manipulator and its optimal trajectories in task space, with and without gravity. Each link has the same mass ($ m $) and length ($ \ell $), and the joint angle of the $ i $-th link is denoted as $ \config_i $. Gravity acts downward. (b) The same optimal trajectories in configuration space, where the potential field is shown as a filled contour. Tissot’s indicatrix (ellipses and cross lines) visualizes the inertia metric, indicating spatial contraction/expansion and the velocity needed for unit kinetic energy. (c) Optimal trajectories in a metric-stretched space, approximating the Riemannian manifold's geometry by scaling basis vectors for uniform kinetic energy representation~\cite{ramasamy_geometry_2019}.}
    \label{fig:TwoLinkOptimal}
\end{figure}
To get a system of equations that determine the optimal trajectory $\config(t)\in \configspace$ in this setting, we shall use the Pontryagin Maximum Principle (PMP).\footnote{Alternatively, one can look directly for critical points of the cost $\costfcn$.}

Following~\cite{balseiro_about_2017,cabrera_optimal_2024}, using the Levi-Civitta connection $\nabla\equiv \nabla^\gmtensor$, the Hamiltonian system $(\hamvarspace,\Omega,\hamfcn)$ coming from PMP has phase space isomorphic to
\begin{multline}
    \hamfcn = \tans^*\configspace\times_Q \tans\configspace \times_Q \tans^*\configspace, \ \Omega = \omega_\nabla, 
    \\ \text{ with variables } \hamdvar,\hamvar \in \tans[\config]^*\configspace, \dconfig\in \tans[\config]^*\configspace,
\end{multline}
where $\omega_\nabla$ is isomorphic to the canonical symplectic form under a diffeomorphism $P\simeq \tans^*(\tans\configspace)$ (see~\cite{balseiro_about_2017}). The Hamiltonian $\hamfcn= \max_{\coninput} (H_{\coninput})$ comes from the family
\begin{equation}
\hamfcn_{\coninput} = - \inmtensor_{ij}{\coninput}^i{\coninput}^j + \hamdvar_i\dconfig^i + \hamvar_i\coninput^i + \hamvar_i\extacc^i, 
\end{equation}
leading to
\begin{equation}
\hamfcn = \frac{1}{4}\inmtensor^{ij}\hamvar_{i}\hamvar_{j} + \hamvar_i\extacc^i + \hamdvar_i\dconfig^i, \ (\coninput_{max})^i = \dfrac{1}{2} \inmtensor^{ij}\hamvar_j.
\end{equation}
Using the formulas for $\omega_\nabla$ and the corresponding Hamiltonian vector fields in~\cite[Prop. 1]{balseiro_about_2017}, we arrive at the following system of equations written in natural coordinates induced by coordinates $(\config^i)$ in $\configspace$,
\begin{proposition}
\label{prop:optconeq}
The optimal trajectory $\config(t) \in \configspace$ for the control system defined by the data $(\configspace,\gmtensor,\inmtensor,\extacc)$ as above, is given by a solution of the following Hamiltonian system: for unknowns $\config(t)\in \configspace$, $\hamdvar(t),\hamvar(t)\in T^*_{\config(t)}\configspace$ and $\dconfig(t)\in \tans[\config(t)] \configspace$,
\begin{align}
    \left(\covderiv{\dconfig}\dconfig\right)^i &= \frac{1}{2} \inmtensor^{ij}\hamvar_j + \extacc^i, \nonumber \\
    \left(\covderiv{\dconfig}^*\hamvar\right)_i &= - \hamdvar_i - \hamvar_j (\partial_{\dconfig^i}\extacc^j), \nonumber \\
    \left(\covderiv{\dconfig}^*\hamdvar\right)_i &= - \frac{1}{4}\tau_i^{jk} \hamvar_j \hamvar_k + \rcurv_{ijk}^l \dconfig^j \dconfig^k \hamvar_l \nonumber \\
    & - \big[ \underbrace{\hamvar_k \left(\covderiv{\vbasis{i}}\extacc\right)^k - \Gamma_{ij}^k \dconfig^j (\partial_{\dconfig^k}\extacc^l)\hamvar_l}_{\text{External force terms}}  \big],
\label{eq:optsplinecoord}
\end{align}
where, as in~\cite[Appendix A]{cabrera_optimal_2024}, $R$ and $\covderiv{}$ are associated with a metric tensor $\gmtensor$, and the tensor $\tau_i^{jk}\equiv \covderiv{\bullet}^* \inmtensor^*$ is given by a covariant derivative of a dual induced tensor,
\begin{equation}
\tau^{jk}_i(\config) = \vbasis{i}\inmtensor^{jk} + \Gamma_{il}^j \inmtensor^{lk} + \Gamma_{il}^k \inmtensor^{jl}.
\end{equation}
\end{proposition}
The proof of \cref{prop:optconeq} follows straightforwardly from~\cite{balseiro_about_2017,cabrera_optimal_2024} keeping track of the extra terms. For the purpose of this paper, we skip the coordinate-invariant version of the equations. The coordinate-invariant approach will be discussed elsewhere.

\section{Geometric Intuition}
\label{sec:geomeaning}
In this section, we describe the geometric meaning of the optimal spline equation. Even though the optimal control equation in \cref{eq:optsplinecoord} is general for the external force, we narrow down the scope for our interest in the gravitational and drag forces. We consider a simple model for a resistive force and include a potential force,
\begin{equation}
    \extacc^i = -\gmtensor^{ij}\left(\vbasis{j}\PE(\config)+\gdtensor_{jk}(\config)\dconfig^k\right), 
\end{equation}
where $\gdtensor$ is a drag tensor containing information about the drag coefficient for each direction with respect to a configuration, $\PE:\configspace \to \euclid$ is the potential energy as a smooth function of a configuration (i.e.,\ a smooth scalar field) on a configuration manifold $\configspace$, and $\vbasis{j}\PE$ is a partial derivative of potential energy along $\config^j$ direction. With this setting, the optimal control system of interest is defined by the data $(\configspace,\gmtensor,\inmtensor,\extforce) \equiv (\configspace,\gmtensor,\inmtensor,\PE,\gdtensor)$.

For a straightforward explanation through this section, we consider the acceleration-based norm cost, which makes the dual inertia metric so that the data we consider is reduced to $(\configspace,\gmtensor,\PE,\gdtensor)$,
\begin{equation}
    \inmtensor = \gmtensor \text { and } \coninput^i = \dfrac{1}{2} \gmtensor^{ij}\hamvar_j.
    \label{eq:assum1}
\end{equation}

\subsection{Inertial Effect}

Consider the situation without the external force and with the acceleration-based cost, the optimal spline equation in \cref{eq:optsplinecoord} reduces to a Riemannian spline equation governed by the Riemannian curvature tensor term,
\begin{equation}
    \left(\covderiv{\dconfig}^2a\right)^i+\rcurv^i_{jkl}\dconfig^j\dconfig^k a^l=0.
    \label{eq:curvaturecond}
\end{equation}
The Riemannian spline equation resembles the geodesic variation (Jacobi) equation~\cite{lee_riemannian_1997}, allowing for a similar geometric interpretation. On flat manifolds with zero curvature, control input $\coninput$ increases or decreases linearly along the trajectory. On manifolds of positive curvature, geodesics tend to converge, and the Riemannian curvature term causes control inputs to point inward, leading to trajectory convergence. Conversely, on manifolds of negative curvature, control inputs point outward, resulting in trajectory divergence.

\subsection{Potential Force Effect}
Consider a conservative dynamical system, which means that the external force only includes a potential force. The equivalent acceleration generated by the potential force is a Riemannian gradient of potential energy with respect to an inertia metric $\gmtensor$: $\grad \equiv \grad[\gmtensor]$ ~\cite{lee_riemannian_1997,bullo_geometric_2005},
\begin{equation}
    -(\grad{}\PE)^i = \gmtensor^{ij}\vbasis{j}\PE = \extacc^i.
    \label{eq:riemgrad}
\end{equation}

\begin{figure}
    \centering
    \includegraphics[width=\linewidth]{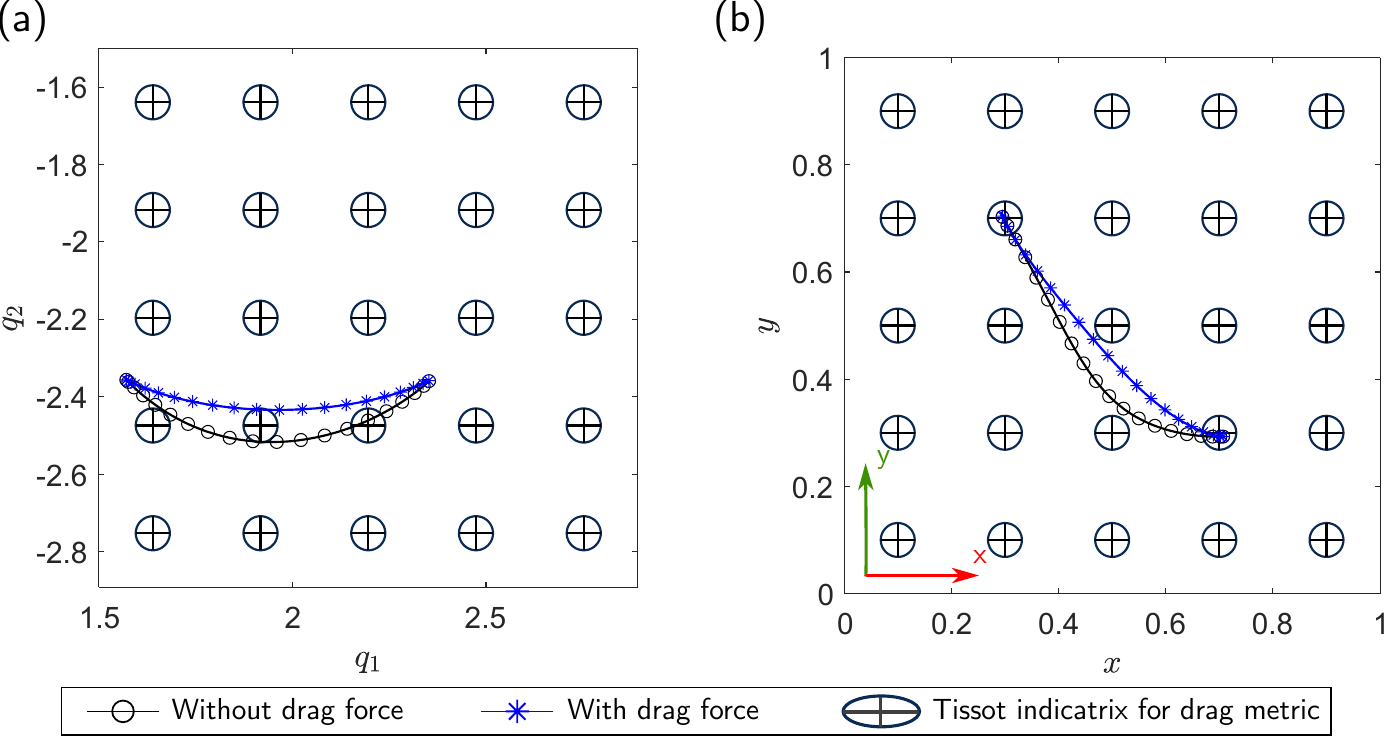}
    \caption{(a) Optimal trajectories of a two-link manipulator with and without joint friction. The black and blue curves represent the trajectories without and with friction, respectively. In configuration space, joint friction follows an Euclidean metric, resulting in circular-shape Tissot indicatrices. The trajectory with friction is closer to a straight path. (b) Optimal trajectories of the same manipulator with an end-effector drag force. The black and blue curves represent trajectories without and with drag. In task space, drag follows an Euclidean metric, yielding circular-shape Tissot indicatrices. The trajectory with drag is closer to a straight path.}
    \label{fig:TwoLinkOptimalDrag}
\end{figure}

\begin{figure*}[ht]
    \centering
    \includegraphics[width=\linewidth]{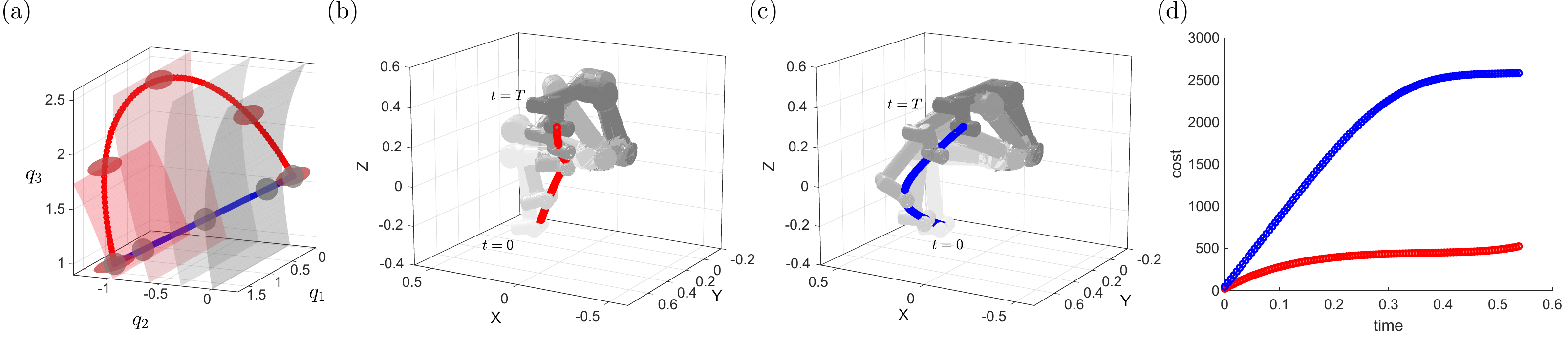}   \caption{(a) Optimal trajectories for a UR5 moving its first three joints under the combinational effect of inertia, gravitation, and drag force. Tissot indicatrices visualize the metric space at given points of the trajectory. The red-colored curve represents an optimal trajectory under the actuation-based cometric, while the blue-colored curve corresponds to the Euclidean metric, equivalent to interpolation between initial and final conditions. The set of surfaces denotes a level set of a potential field. (b) An optimal trajectory of the UR5 manipulator under the actuation-based cometric in task space. (c) An optimal trajectory under the Euclidean metric or interpolation between initial and final conditions in task space. In (b) and (c), the manipulator transitions from white to gray, representing its initial and final configurations. (d) A time versus acceleration norm cost, defined as $\int_{t_0}^{t_f}\norm{\coninput}^2_{\gmtensor}dt$.}
    \label{fig:UR5Example}
\end{figure*}
A Riemannian Hessian of a scalar field (e.g., a potential field) associated with an inertia metric $\gmtensor$ describes a second-order derivative effect of a scalar field on the manifold: $\hess \equiv \hess[\gmtensor]$~\cite{lee_riemannian_1997,bullo_geometric_2005},
\begin{equation}
    (\hess\PE)_{ij} = \vbasis{i}\vbasis{j}\PE - \ctfl{k}{i}{j}\vbasis{k}\PE = \gmtensor_{jk}(\covderiv{\vbasis{i}}\grad{}\PE)^k.
    \label{eq:riemhess}
\end{equation}
A Riemannian Hessian of a potential field describes the local curvature of a function with respect to tangent vectors on Riemannian manifolds.

The potential force term in \cref{eq:optsplinecoord} can be simplified using \cref{eq:riemgrad,eq:riemhess}, yielding a Riemannian Hessian of the potential energy:
\begin{multline}    
\hamvar_k \left(\covderiv{\vbasis{i}}\extacc\right)^k = -\hamvar_k\left(\covderiv{\vbasis{i}}\grad{}\PE\right)^k \\ = -\gmtensor^{jk}\hamvar_k(\hess\PE)_{ij} = -2\coninput^j(\hess\PE)_{ij}.
\end{multline}
With the assumption \cref{eq:assum1}, the potential energy term corresponds to the contraction of the potential Hessian with the control input. Due to the sign difference in the definition of potential force, a positive local curvature of the potential field directs the control input outward. From a mechanical perspective, the controller leverages gravitational force to minimize control effort. Consequently, the optimal trajectory tends to follow paths where the control input aligns with changes in potential force, while avoiding misaligned directions. 
 
% The potential energy term in optimal spline equation \cref{eq:optsplinecoord} is equivalent to the vector contracting potential Hessian with a control input $\coninput$. Because of the sign difference from the definition of the potential force, a positive local curvature of potential fields makes the control input points outward. From the viewpoint of mechanics, the controller uses the gravity force to reduce the use of the control input as much as possible. Thus, the optimal trajectory prefers the route where the control input aligns with the direction changing in a potential force and avoids the misaligned direction.
 
\textit{Examples}: 
As illustrated in \cref{fig:TwoLinkOptimal}, we compare two optimal trajectories for a two-link manipulator with and without gravity to evaluate its effect. The inertia tensor is derived by pulling back each link's individual inertia tensor through its Jacobian. A detailed derivation of a mass matrix and a potential field can be found in~\cite{lynch_modern_2017,hatton_geometry_2022}. For this example, we use a torque norm as the cost functional, with an actuation-based cometric tensor that reduces to the identity matrix due to its physical configurations. Two scenarios of optimal control are considered. The initial and final configuration conditions of each scenario are
\begin{subequations}
\begin{align}
    1) \ & q_0 = \transpose{\begin{bmatrix}\frac{1}{2}\pi&-\frac{3}{4}\pi\end{bmatrix}}, & & q_T = \transpose{\begin{bmatrix}\frac{3}{4}\pi&-\frac{3}{4}\pi\end{bmatrix}}, \\
    2) \ & q_0 = \transpose{\begin{bmatrix}2& -2.3\end{bmatrix}}, & & q_T = \transpose{\begin{bmatrix}2 & -1.7\end{bmatrix}}.
\end{align}    
    \label{eq:initfinalconds}
\end{subequations}
The initial and final velocity conditions of both scenarios are zero velocities.

Roughly speaking, the left-upper region of the potential field in \cref{fig:TwoLinkOptimal}(b) has a positive curvature, and the left-lower region has a negative curvature. Thus, optimal trajectories under the gravity effect are more distorted toward the negative curvature of the potential field than those without the gravity effect. Especially, if we set the initial and final conditions so that the optimal trajectories are aligned with the direction of a curvature effect (e.g., a vertical trajectory in \cref{fig:TwoLinkOptimal}(b)), a phase and an end time of trajectories with a gravity effect are different with those without a gravity effect.

\subsection{Resistive Force Effect}
Unlike a potential force, a resistive force such as friction or drag force dissipates energy in mechanical systems. Thus, we can expect a trajectory with a short path length under the drag geometry to be advantageous. In other words, adding the drag effect to the optimal control would make the shape of an optimal trajectory closer to a geodesic shape on the drag manifold. 

With drag effects, the cost functional can be decomposed into three distinct terms by combining \cref{eq:coneq,eq:concost}:
\begin{multline}
    \norm{\coninput}_{\gmtensor}^2 = \gmtensor_{ij}\coninput^i\coninput^j \\ = \gmtensor_{ij}(\covderiv{\dconfig}\dconfig)^i(\covderiv{\dconfig}\dconfig)^j + \gdtensor'_{ij}\dconfig^i\dconfig^j - 2(\force_{D})_{j}(\covderiv{\dconfig}\dconfig)^j.
\end{multline}
where $(\force_{D})_{j} = \gdtensor_{ij}\dconfig^i$ is a drag force covector, and $\gdtensor' = \transpose{\gdtensor}\gmtensor\gdtensor$ is an induced drag metric tensor. \begin{itemize}
    \item The first term represents the covariant acceleration norm, primarily influenced by the inertia manifold.  
    \item The second term corresponds to the drag force norm associated with the inertia metric, or equivalently, the velocity norm on the drag manifold. Minimizing the integral of this norm favors shorter paths on the drag manifold and a constant-speed profile~\cite{kelly_geometric_1995, ramasamy_geometry_2020}. 
    \item The third term captures the coupling effect, given by the product of covariant acceleration and drag forces. 
    \item Without drag effects, the second and third terms vanish, reducing the formulation to the classical inertia-dominated case. 
\end{itemize}

\textit{Examples}: As illustrated in \cref{fig:TwoLinkOptimalDrag}, we apply optimal control to achieve four different scenarios of optimal trajectories for a two-link manipulator with and without two types of drag force: 1) a joint friction and 2) a drag force applying to the end-effector. We assume that the drag coefficient is the same in each direction and use a viscous friction model for simplicity. The joint friction tensor is identical to the Euclidean metric. The end-effector drag tensor can be constructed by pulling back the inertia tensor of the end link through the kinematic Jacobian. A detailed derivation of a drag tensor can be found in~\cite{ramasamy_geometry_2019}. We considered the same as the first scenario in \eqref{eq:initfinalconds}.

\cref{fig:TwoLinkOptimalDrag}(a) illustrates the comparison between the trajectory with and without the joint friction. The optimal trajectory without joint friction is distorted because of the curvature effect of the inertia manifolds. On the other hand, the optimal trajectory with joint friction is distorted by combining the curvature effect with the straightening effect of the drag manifold. \cref{fig:TwoLinkOptimalDrag}(b) illustrates the comparison between the trajectory with and without the drag force applied to the end-effector. In task space, the drag coefficient in each direction is identical. Similarly, the optimal trajectory with an end-tip drag force is more straight than the optimal trajectory without an end-tip drag force.

\section{Results}
\label{sec:results}
We applied the methods in \cref{eq:optsplinecoord} to construct an optimal control scheme for a UR5 manipulator. The initial and final configuration conditions are $q_0 = \transpose{\left[0.1, 0.1, \pi/2 \right]}$, $q_T = \transpose{\left[\pi/2, -1, 1\right]}$ and fixed the remaining joint values as $0.1$. The initial and final velocity conditions are zero velocities. Note that we only give the command to the first three joints starting from the base joint to decrease computational cost and visualize the trajectory, even though the method can be applied to high-dimensional systems. The dynamical system of UR5 includes the gravity and joint friction forces. The mass and linkage length information can be found in~\cite{kebria_kinematic_2016}. We assume a linear friction model for simplicity. To visualize the robots, we use \verb|MATLAB|'s \verb|Robotics System Toolbox|. To solve the boundary value problems in this paper, we use the multiple shooting method based on the \verb|CasADi|'s automatic differentiation~\cite{andersson_casadi_2019}.

As illustrated in \cref{fig:UR5Example}, we compare two optimal trajectories for the UR5 manipulator under the same initial and final conditions, accounting for joint friction and gravitational effects. The Euclidean metric trajectory interpolates between the start and end states, while the actuation-based cometric trajectory accounts for inertia, resulting in a path that may appear longer in joint space but is more efficient. \cref{fig:UR5Example}(b-c) illustrates the differences between the actuation-optimal and Euclidean metric trajectories in task space. The actuation-optimal trajectory initially folds the arm to minimize inertia effects and then stretches it to reach the final configuration, where inertia has less impact. As illustrated in \cref{fig:UR5Example}(d), the cost of the actuation-optimal trajectory increases moderately over time compared to the Euclidean metric trajectory and represents a more efficient route overall so that the control input (or the trajectory) aligns with the shorter axis of the Tissot ellipsoids.

\section{CONCLUSIONS}
In this paper, we have presented a geometric framework for understanding optimal motions in mechanical systems, considering the effects of inertia and external forces (e.g., drag and gravitational forces). By employing Riemannian geometry to describe these systems, we gained insights into how these external forces influence optimal trajectories. We derived the optimal control equations based on the Pontryagin Maximum Principle~\cite{balseiro_about_2017,cabrera_optimal_2024} and demonstrated their application to two-link and UR5 robotic manipulators, highlighting the curvature effects of each manifold and the path-shortening influence of resistive forces. This approach not only enhances our understanding of trajectory optimization but also provides a foundation for applying Riemannian variational calculus to a wide range of robotic systems (e.g., locomoting systems~\cite{choi_optimal_2022,yang_geometric_2023}) and mechanical applications. Future work could explore the integration of additional forces and constraints (e.g., nonholonomic constraints~\cite{hussein_optimal_2008}), further broadening the applicability of this geometric approach to more complex systems.

\addtolength{\textheight}{-10.5cm}   % This command serves to balance the column lengths

%%%%%%%%%%%%%%%%%%%%%%%%%%%%%%%%%%%%%%%%%%%%%%%%%%%%%%%%%%%%%%%%%%%%%%%%%%%%%%%

\bibliographystyle{ieeetr}
\bibliography{ref}

%%%%%%%%%%%%%%%%%%%%%%%%%%%%%%%%%%%%%%%%%%%%%%%%%%%%%%%%%%%%%%%%%%%%%%%%%%%%%%%%

\end{document}